\documentclass[letterpaper, 10 pt, conference]{./support/ieeeconf}
\usepackage{times}
\usepackage[pdftex]{graphicx}
\usepackage{amsmath,amssymb,amsopn,amstext,amsfonts}
\usepackage{cancel}
\usepackage[space]{cite}
\usepackage{pdfsync}
\usepackage{balance}
\usepackage{color}
\usepackage{mathtools}
\usepackage{algpseudocode}
\usepackage{algorithm} %\usepackage[ruled,vlined,linesnumbered]{algorithm2e}
\usepackage{bm}

\usepackage{diagbox}
\usepackage{float}
\usepackage{epstopdf}
\usepackage{pifont}
\usepackage{fixltx2e}
\usepackage{amsmath}
\usepackage{multirow}
\usepackage{url}
\usepackage{verbatim}
\makeatletter
\let\NAT@parse\undefined
\makeatother
\usepackage[linkcolor=black,citecolor=black,urlcolor=black,colorlinks=TRUE]{hyperref}
\usepackage{booktabs}
\usepackage{threeparttable}
\usepackage{subfigure}
\usepackage{caption}

\graphicspath{{./figures/}}
\DeclareGraphicsExtensions{.png,.jpg,.eps,.pdf}
\IEEEoverridecommandlockouts
\title{\textbf{VID-Fusion: Robust Visual-Inertial-Dynamics Odometry \\ for Accurate External Force Estimation}}
\author{Ziming Ding\textsuperscript{1,2,3}, Tiankai Yang\textsuperscript{1,2,3}, Kunyi Zhang\textsuperscript{1,2,3}, Chao Xu\textsuperscript{1,2}, and Fei Gao\textsuperscript{1,2}
	\thanks{This work was supported by National Natural Science Foundation of
		China under Grant 62003299.}
	\thanks{\textsuperscript{1}State Key Laboratory of Industrial Control Technology, Institute of Cyber-Systems and Control, Zhejiang University, Hangzhou 310027, China.}
	\thanks{\textsuperscript{2}Huzhou Institute, Zhejiang University, Huzhou 313000, China.}
	\thanks{\textsuperscript{3}3National Engineering Research Center for Industrial Automation (Ningbo Institute), Ningbo 315000, China.}
	\thanks{E-mail: {\small \{zm\_ding, and fgaoaa\}@zju.edu.cn}}}

\begin{document}
	
	\maketitle
	\thispagestyle{empty}
	\pagestyle{empty}
	
	\begin{abstract}
	Recently, quadrotors are gaining significant attention in aerial transportation and delivery.
In these scenarios, an accurate estimation of the external force is as essential as the six degree-of-freedom (DoF) pose since it is of vital importance for planning and control of the vehicle.
To this end, we propose a tightly-coupled Visual-Inertial-Dynamics (VID) system that simultaneously estimates the external force applied to the quadrotor along with the six DoF pose.
Our method builds on the state-of-the-art optimization-based Visual-Inertial system~\cite{qin2018vins}, with a novel deduction of the dynamics and external force factor extended from VIMO~\cite{nisar2019vimo}.
Utilizing the proposed dynamics and external force factor, our estimator robustly and accurately estimates the external force even when it varies widely.
Moreover, since we explicitly consider the influence of the external force, when compared with VIMO~\cite{nisar2019vimo} and VINS-Mono~\cite{qin2018vins}, our method shows comparable and superior pose accuracy, even when the external force ranges from neglectable to significant.
The robustness and effectiveness of the proposed method are validated by extensive real-world experiments and application scenario simulation.
We will release an open-source package of this method along with datasets with ground truth force measurements for the reference of the community.
	\end{abstract}
	
	\IEEEpeerreviewmaketitle
	
	\section{Introduction}
	\label{sec:introduction}
	Micro Aerial Vehicles (MAVs), especially quadrotors, have long been paying expectancy in package delivery, aerial manipulation, and collaborative transportation.
In these applications, quadrotors are required to carry heavy payloads or suspended cargoes, which apply external forces periodically to the drone.
These forces are un-neglectable, thus causing large disturbances and significantly harming the flight of the quadrotor.
To compensate for the disturbances, the controller and planner of a quadrotor have to precisely know the magnitude and direction of the external force in advance.
Therefore, it is necessary to design a state estimator, which takes external forces into explicit consideration.
	
	Recently, the Visual Inertial Odometry (VIO)~\cite{leutenegger2015keyframe,qin2018vins} system has shown its reliable capability to provide robust and accurate state estimation of a quadrotor, even within the minimalist hardware configuration, such as a monocular or stereo camera with an inertial measurement unit (IMU).
To combine the external disturbance force estimation into a VIO system, VIMO~\cite{nisar2019vimo} formulates a model-based VIO in a tightly coupled manner~\cite{qin2018vins} and opens the door to simultaneously estimate motion and external forces with a VIO system.
By considering the dynamics model of the quadrotor as a new information source, VIMO improves the accuracy of pose estimation and utilizes the dynamics to provide a clue for the external force estimation.
VIMO~\cite{nisar2019vimo} jointly estimates the motion and forces by modeling the unknown external forces as zero-mean Gaussian values, assuming the external force estimated is near zero or only occurs in a short period.
Obviously, this assumption only holds for narrow scenarios and results in severe unstability or even failure to pose estimation when the drone receives large or continuous forces.
	
\begin{figure}[t]
		\centering
		\includegraphics[width=1.0\columnwidth]{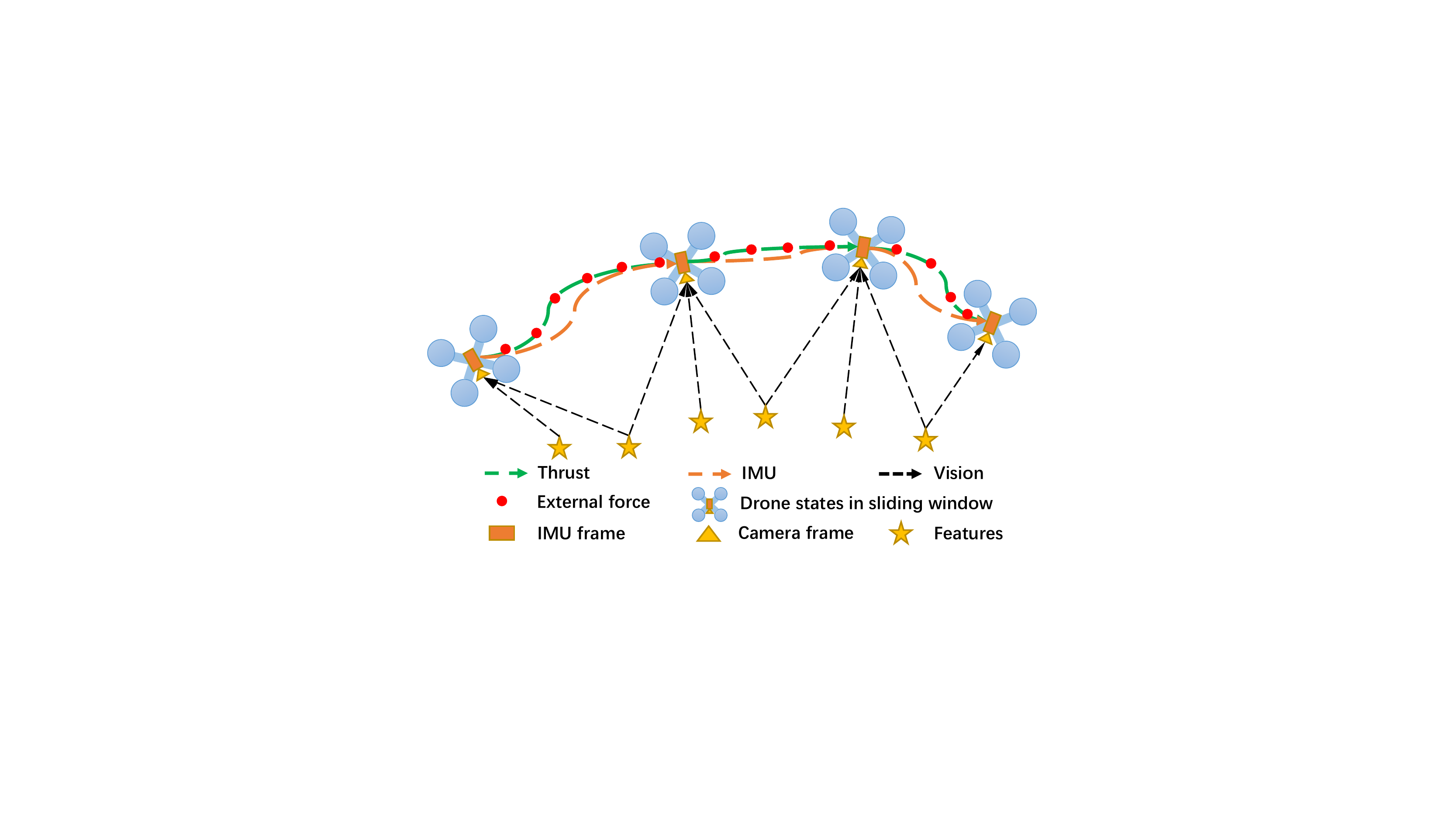}
		%\vspace{-0.5cm}
		\captionsetup{font={small}}
		\caption{This figure illustrates the relationship among thrust, external force observation, and IMU measurements in the sliding window. The integration from a combination of thrust and external force, which is equal to IMU measurements integration ideally, can both represent the movement between two continuous frames of drone states. Therefore, the thrust and external force will be added into the sliding window as an extra factor, formulating the model-based visual-inertial force estimation framework.
		\label{fig:slidingwindow}}
		\vspace{-1.0cm}
	\end{figure}

To accurately model and measure the effects of external forces, it is necessary to incorporate the quantification of external force into a VIO system.
As illustrated in Fig.~\ref{fig:slidingwindow}, the external force is roughly equal to the discrepancy between motions measured from IMU and thrust measurements without considering sensor noise characteristics.
In this paper, instead of using the zero-mean Gaussian assumption, we provide a rough force observation by comparing IMU and thrust.
We derive an external force preintegration term to represent the average force between consecutive camera frames and then combine the thrust preintegration proposed in~\cite{nisar2019vimo} to reformulate the dynamics residual.
Finally, the stacked residual is jointly minimized by using nonlinear optimization within the sliding window, as shown in Fig.~\ref{fig:slidingwindow}.

	In this paper, we improve the force estimation based on the model-based VIO~\cite{nisar2019vimo} and propose an optimization-based Visual-Inertial-Dynamics (VID) framework extended from \cite{leutenegger2015keyframe,qin2018vins,forster2016manifold}. 
	Compared with VIMO~\cite{nisar2019vimo} and VINS-Mono~\cite{qin2018vins}, the proposed estimator demonstrates superior pose and force estimation accuracy and attains robustness even when the external force varies widely. 
	
To summarize our contributions:
\begin{itemize}
\item We derive the external force preintegration and propose state, covariance propagation, and force bias correction for discrete-time processing.
\item We incorporate the external force preintegration into the state-of-the-art VIO framework and present a complete, robust, tightly-coupled Visual-Inertial-Dynamics state estimator.
\item We demonstrate robust and accurate external force and pose estimation onboard the quadrotor and present a new dataset with ground truth measurements.
\end{itemize}
	
The rest of the paper is organized as follows. Related works are shown in Sec.II. We describe the proposed framework in detail in Sec.III. Following several experiments in Sec.IV, we show the performance of algorithms in this article. Finally, we give relevant conclusions on our algorithm and future work in Sec.V.
	
	\section{Related works}
	\label{sec:related_works}
	\subsection{System model and identification}
	
	For better external force estimation, we must distinctly comprehend the source of the driven force of MAVs.~\cite{Online2019Wuest} describes an online Kalman filter-based method to identify a quadrotor system, especially geometrical and inertial parameters. However, dynamics parameters such as the thrust coefficient still need to be measured by experiments. In~\cite{2017Introduction}, the authors give us almost all the details corresponding to a multi-copter system, but it demands offline experiments with a specialized device to obtain the relation between rpm and aerodynamic force, including thrust.
	
	\subsection{Motion state estimation}
	Before disturbance estimation, the motion state is the hot spot in the relevant field. Especially, visual-inertial odometry has achieved a lot of excellent achievements and extensive applications. MSCKF~\cite{li2013high}, VINS-Mono~\cite{qin2018vins}, Open-VINS~\cite{geneva2020openvins} and so on are some brilliant open-source works. All algorithms above work well in most conditions due to effective, robust and online features, but they neglect the dynamics of the robots, focusing only on the ego-motion state estimation with as little as possible sensors.
	So in complex environments~\cite{Tagliabue2020TouchTW,2020Robust,Ji2020CMPCCCM} with contacts, interactions and disturbances, we can not achieve an ideal goal only with an unreliable setting value of disturbance.
	
	\subsection{External force estimation}
	Roughly, we can obtain the external force directly from the collective thrust and acceleration measurements~\cite{Teodor2014A}. A nonlinear observer such as~\cite{Yuksel2014A,2014Impedance,Disturbance2018Kim} is designed for disturbance observation, but they require good dynamics models and enough measurements.
	The filter-based method is usually used for denoising.
	~\cite{2016Unscented,2013Admittance,2016Collaborative,2014Model} use a loosely-coupled filter-based method to estimate a disturbance but neglecting the interaction of different sensor data, resulting in a draft or bias of the estimator.
	VIMO~\cite{nisar2019vimo} bridges the gap between motion estimation and external force estimation within an optimization-based VIO system, but the unknown external force is modeled as a zero-mean Gaussian. So the estimator will be highly affected and even leads to failure of the system by large or continuous forces. 
	
	\section{Methodology}
	\label{sec:methodology}
	
	In this section, we will describe the proposed Visual-Inertial-Dynamics framework in detail. Fig.~\ref{fig:axis} shows the definition of the coordinate system, where body frame, IMU frame and the geometric frame are assumed coincided, denoted by ${[\cdot]}^{b}$. Otherwise,  ${[\cdot]}^{w}$ denotes the world frame, and $\hat{[~]}$ represents the value with measurement.

	\subsection{External Force Preintegration}
	We formulate our external force preintegration inspired by acceleration preintegration in~\cite{shen2015tightly} and~\cite{forster2016manifold}, and extend preintegration of dynamics factor in~\cite{nisar2019vimo} to be a dynamics and external force preintegration factor. 
	
	\subsubsection{External Force Quantification}
	The quantification of the external force exerted on a drone needs to combine accelerometer data from inertial measurement unit (IMU) and the rotor thrust from the rotor rotating speed measurement unit (RMU), in which the raw acceleration $\boldsymbol{\hat{a}}^b$ and the mass normalized thrust acceleration $\boldsymbol{\hat{T}}^b$ are modeled as:
	\begin{equation}
	\begin{aligned}
	\boldsymbol{\hat{a}}^b &= \boldsymbol{a}^b + \boldsymbol{b}_{a} + \mathbf{R}^{b}_w \boldsymbol{\textsl{g}}^w + \boldsymbol{n}_a \\
	\boldsymbol{\hat{T}}^b &= \frac{1}{m}\sum_{i=1}^{n_{rotor}}\big(\tau_{f_i} \omega_{i}^2) + \boldsymbol{n}_T, \\
	\end{aligned} 
	\end{equation}
	where the noise of acceleration and thrust measurements are assumed as Gaussian white noise:
	\begin{equation}
	\begin{aligned}
	\boldsymbol{n}_a \sim\mathcal{N}(\boldsymbol{0},\boldsymbol{\sigma}^2_{a}),~~~ \boldsymbol{n}_T \sim\mathcal{N}(\boldsymbol{0},\boldsymbol{\sigma}^2_{T}).
	\end{aligned} 
	\end{equation}
	$\boldsymbol{a}^b$ is true value of acceleration in IMU frame, and $\boldsymbol{b}_{a}$ is the acceleration measurement bias, whose derivative is also assumed as Gaussian white noise with
	\begin{equation}
	\begin{aligned}
	\dot{\boldsymbol{b}}_{a}=\boldsymbol{n}_{b_a},~~~ \boldsymbol{n}_{b_a} \sim\mathcal{N}(\boldsymbol{0},\boldsymbol{\sigma}^2_{b_a}). 
	\end{aligned} 
	\end{equation}
	$\mathbf{R}^{b}_w$ is the rotation matrix from the world frame to body frame, requiring the rotation evolution provided by IMU rotation model  $\dot{\boldsymbol{q}}^w_{b}=\boldsymbol{q}^w_{b}\otimes[0,\boldsymbol{\omega}^b]^T$, where $\boldsymbol{q}^w_{b}$ is the quaternion for the rotation, and $\otimes$ is the quaternion multiplication and $\boldsymbol{\omega}^b$ is the body angular velocity, and the gyroscope measurements with noise and bias from IMU is
	\begin{equation}
	\begin{aligned}
	\boldsymbol{\hat{\omega}}^b=\boldsymbol{\omega}^b+\boldsymbol{b}_{\omega}+\boldsymbol{n}_{\omega},
	\end{aligned} 
	\end{equation} 
	$\boldsymbol{n}_{\omega}$ and the derivative of $\boldsymbol{b}_{\omega}$ are also considered as Gaussian white noise:
	\begin{equation}
	\begin{aligned}
	\dot{\boldsymbol{b}}_{\omega}=\boldsymbol{n}_{b_\omega},~~ \boldsymbol{n}_{b_\omega} \sim\mathcal{N}(\boldsymbol{0},\boldsymbol{\sigma}^2_{b_\omega}),~~
	\boldsymbol{n}_{\omega} \sim\mathcal{N}(\boldsymbol{0},\boldsymbol{\sigma}^2_{\omega}).
	\end{aligned} 
	\end{equation}
	\begin{figure}[t]
		\captionsetup{font={small}}
		\centering
		\includegraphics[width=1\columnwidth]{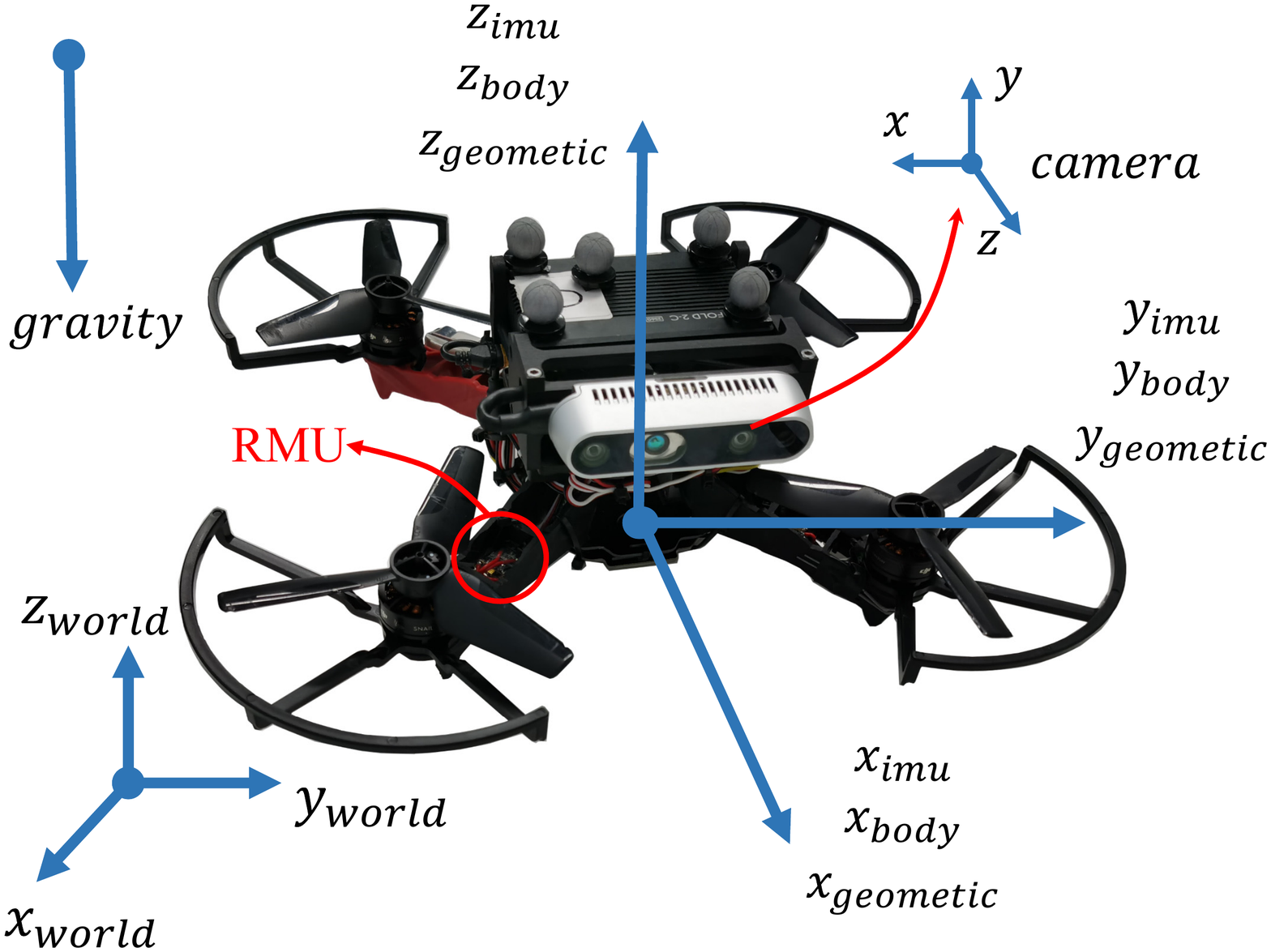}
%		\vspace{-0.5cm}
		\captionsetup{font={small}}
		\caption{The frames defined in our system, where body frame, IMU frame and the geometric frame are assumed coincided.  \label{fig:axis}}
	\vspace{-0.5cm}
	\end{figure}
	The gravity vector is \mbox{$\boldsymbol{\textsl{g}}^w=[0,0,-9.79]m/s^2$}. Besides, $m$ stands for mass, and $n_{rotor}$ stands for the number of rotors, and $\tau_{f_i}$ is thrust coefficient for $i$th rotor of any arbitrary multi-rotors, and $\omega_i$ is the $i$th rotor rotation speed from RMU. More details and system identification about thrust coefficient will be discussed in \ref{subsec:model_identification}.
	Here, we assume that the IMU frame is the body frame, and the geometric center of the drone is coincident with the drone centroid and IMU center, as shown in Fig.~\ref{fig:axis}. They are all called the body frame in our framework. We consider external force as a resultant force except for rotor thrust and aircraft gravity, acting on the mass center. So we can derive the relationship between $\boldsymbol{\hat{a}}^b$ and $\boldsymbol{\hat{T}}^b$:
	\begin{equation}
	\begin{aligned}
	\boldsymbol{a}^b  
	&= \boldsymbol{\hat{T}}^b + \boldsymbol{f}^b_{ex} - \mathbf{R}^{b}_w \boldsymbol{\textsl{g}}^w - \boldsymbol{n}_T \\
	&= \boldsymbol{\hat{a}}^b - \boldsymbol{b}_{a} - \mathbf{R}^{b}_w \boldsymbol{\textsl{g}}^w - \boldsymbol{n}_a.
	\end{aligned} 
	\end{equation}
	$\boldsymbol{f}^b_{ex}$ represents the external force observation, combined by rotor drag and other force. And now we can measure the external force with simple operation as follow:  
	\begin{equation}
	\label{equation:force_measurenment}
	\begin{aligned}
	\boldsymbol{f}^b_{ex} = \big(\boldsymbol{\hat{a}}^b - \boldsymbol{b}_{a} - \boldsymbol{\hat{T}}^b \big) - \big(\boldsymbol{n}_a - \boldsymbol{n}_T\big).
	\end{aligned} 
	\end{equation}

	\subsubsection{External Force Preintegration}
	The external force $\boldsymbol{f}^b_{ex}$ calculated from (\ref{equation:force_measurenment}) have the same dimension as acceleration expressed in body frame, so we can use the acceleration measurement preintegration method to integrate external force between keyframes $b_k$ and $b_{k+1}$ and represent it at frame $b_k$:
	\begin{align}
	\boldsymbol{\hat{J}}^{b_k}_{b_{k+1}} &= \int_{t_k}^{t_{k+1}} \mathbf{R}^{b_k}_{b_\tau} \boldsymbol{f}^{b_\tau}_{ex} d\tau \\
	\boldsymbol{\hat{\mathcal{F}}}^{b_k}_{b_{k+1}} &= \frac{1}{\Delta t_k}\boldsymbol{\hat{J}}^{b_k}_{b_{k+1}}.
	\end{align} 
	$\boldsymbol{\hat{J}}^{b_k}_{b_{k+1}}$ represents the integration of external force over time interval $\Delta t_k=t_{k+1}-t_k$ referencing frame $b_k$. Then we use $\boldsymbol{\hat{\mathcal{F}}}^{b_k}_{b_{k+1}}$ to represent the average force between frames $b_k$ and $b_{k+1}$. During $[t_k,t_{k+1}]$, $\boldsymbol{\hat{\mathcal{F}}}^{b_k}_{b_{k+1}}$ can be considered as a constant force exerting in $b_k$ frame. $\mathbf{R}^{b_k}_{b_\tau}$ is provided by the integration of rotation $\boldsymbol{\hat{\gamma}}{_{b_\tau}^{b_{k}}}$, which will be mentioned in (\ref{equation:propagation}).
	
	We add external force preintegration derived above into preintegration of dynamics factor in~\cite{nisar2019vimo}, and adjust to three preintegration terms containing measurements: $\boldsymbol{\hat{\alpha}}{_{b_{k+1}}^{b_{k}}}$, $\boldsymbol{\hat{\beta}}{_{b_{k+1}}^{b_{k}}}$, $\boldsymbol{\hat{\mathcal{F}}}^{b_k}_{b_{k+1}}$. These terms are integrated from IMU and thrust measurements expressed in frame $b_k$, which are independent of the world frame. 
	\begin{equation}
	\label{eq:preintegration}
	\begin{aligned}
	\boldsymbol{\hat{\alpha}}{_{b_{k+1}}^{b_{k}}} &= \iint_{t{_k}}^{t_{k+1}} \mathbf{R}{^{b_{k}}_{b{_\tau}}} \Big(\boldsymbol{\hat{T}}^{b_\tau}-\boldsymbol{n}_T \Big) d\tau{^2} \\
	\boldsymbol{\hat{\beta}}{_{b_{k+1}}^{b_{k}}}  &= \int_{t{_k}}^{t_{k+1}} \mathbf{R}{^{b_{k}}_{b{_\tau}}}
	\Big(\boldsymbol{\hat{T}}^{b_\tau}-\boldsymbol{n}_T \Big) d\tau     \\
	\boldsymbol{\hat{\mathcal{F}}}^{b_k}_{b_{k+1}} &= \frac{1}{\Delta t_k} \int_{t_k}^{t_{k+1}} \mathbf{R}^{b_k}_{b_\tau} \boldsymbol{f}^{b_\tau}_{ex} d\tau .
	\end{aligned}
	\end{equation}
	$\boldsymbol{\hat{\alpha}}{_{b_{k+1}}^{b_{k}}}$ represents the contribution to distance by the thrust during $[t_k,t_{k+1}]$,  $\boldsymbol{\hat{\beta}}{_{b_{k+1}}^{b_{k}}}$ represents the velocity contributed by thrust, and $\boldsymbol{\hat{\mathcal{F}}}^{b_k}_{b_{k+1}}$ represents the average external force in the period.

	\subsubsection{External Force Propagation}
	As for discrete-time processing, we employ first-order hold numerical integration method as done in~\cite{qin2018vins}. In the period of $[t_k,t_{k+1}]$, every initial value $\boldsymbol{\hat{\alpha}}{_{b_{k}}^{b_{k}}}$,$\boldsymbol{\hat{\beta}}{_{b_{k}}^{b_{k}}}$of propagation at $t_k$ is set to $\boldsymbol{0}$. At the period of $[t_k,t_{k+1})$, the external force $\boldsymbol{\hat{\mathcal{F}}}^{b_k}_{b_{k}}$ propagates in the same way as $\boldsymbol{\hat{\alpha}}{_{b_{k}}^{b_{k}}}$ and $\boldsymbol{\hat{\beta}}{_{b_{k}}^{b_{k}}}$. At the moment of $t_{k+1}$, $\boldsymbol{\hat{\mathcal{F}}}^{b_k}_{b_{k+1}}$ will be divided by $\Delta t_k$, calculating the average force in the time interval  $[t_k,t_{k+1}]$. There is a requirement of rotation matrix $\mathbf{R}^{b_k}_{i}$ during propagation, so quaternion preintegration $\boldsymbol{\hat{\gamma}}{^{b_k}_{i}}$ is employed to propagate from frame $b_k$ to current timestamp $t_i$ of IMU measurements with an initial state of identity quaternion. During the propagation period $[t_k,t_{k+1}]$, the bias of acceleration and gyroscope measurements are considered as constant value, which are the estimated result at time $t_k$, $\boldsymbol{b}_{a_k}$ and $\boldsymbol{b}_{\omega_k}$. The propagation is presented by  
	\begin{equation}
	\label{equation:propagation}
	\begin{aligned}
	\boldsymbol{\hat{\alpha}}{_{i+1}^{b_{k}}} &= \boldsymbol{\hat{\alpha}}{_{i}^{b_{k}}} + \boldsymbol{\hat{\beta}}^{b_k}_i \delta t_i
	+ \frac{1}{2}\mathbf{R}^{b_k}_{i} \boldsymbol{\hat{T}}^{b_i} \delta t{_i^2} \\
	\boldsymbol{\hat{\beta}}{_{i+1}^{b_{k}}}
	&=\boldsymbol{\hat{\beta}}{_{i}^{b_{k}}} + \mathbf{R}^{b_k}_{i} \boldsymbol{\hat{T}}^{b_i} \delta t_i \\
	\boldsymbol{\hat{\mathcal{F}}}^{b_k}_{i+1}
	&= \boldsymbol{\hat{\mathcal{F}}}^{b_k}_{i} + \mathbf{R}^{b_k}_{i} \big(\boldsymbol{\hat{a}}^{b_i} - \boldsymbol{b}_{a_k} - \boldsymbol{\hat{T}}^{b_i} \big) \delta t_i \\
	\boldsymbol{\hat{\gamma}}{^{b_k}_{i+1}}
	&=\boldsymbol{\hat{\gamma}}{_{i}^{b_{k}}} \otimes
	\begin{bmatrix}
	1\\
	\frac{1}{2}(\boldsymbol{\omega}{^{b_i}}-\boldsymbol{b}{_{\omega_k}})\delta {t_i}
	\end{bmatrix},
	\end{aligned}
	\end{equation}
	where $\delta t_i$ is time interval between two continuous IMU measurements $i$ and $i+1$. Here, $\boldsymbol{\hat{T}}^{b_i}$ updates at the rate of RMU, while $\boldsymbol{\hat{a}}^{b_i}$ and $\boldsymbol{\omega}{^{b_i}}$ at IMU rate. Propagation runs at the faster measurement rate. When the faster measurement update, the slower one will keep the latest measurement, and both of them will be propagated together.    
	
	As for propagation of the discrete-time dynamics during $[t_k,t_{k+1}]$, the error state \mbox{$\delta\boldsymbol{z}=[\delta\boldsymbol{\alpha}^T~\delta\boldsymbol{\beta}^T~\delta\boldsymbol{\mathcal{F}}^T~\delta\boldsymbol{b}_a^T~ \delta\boldsymbol{b}_{\omega}^T]^T$} and the state covariance matrix $\mathbf{P}^{b_k}_{i}$ will be propagated by a linearized model with initialization of $\delta\boldsymbol{z}^{b_k}_{b_k}=\boldsymbol{0}, \mathbf{P}^{b_k}_{b_k}=\mathbf{0}$, while the first-order Jacobian matrix $\mathbf{J}_i$ of $\boldsymbol{z}^{b_k}_i$ with respect to $\boldsymbol{z}^{b_k}_{b_k}$ can be also propagated with initialization of $\mathbf{J}_{b_k}=\mathbf{I}$:
	\begin{align}
	\delta\boldsymbol{z}^{b_k}_{i+1} &= \mathbf{F}_i \delta\boldsymbol{z}^{b_k}_{i} + \mathbf{G}_i\boldsymbol{n},  \\
	\label{eq:covarance_propagation}
	\mathbf{P}^{b_k}_{i+1} &= \mathbf{F}_i \mathbf{P}^{b_k}_{i+1} \mathbf{F}_i^T + \mathbf{G}_i \mathbf{Q} \mathbf{G}_i^T, \\
	\mathbf{J}_{i+1} &= \mathbf{F}_i \mathbf{J}_{i},
	\end{align}
	where $\mathbf{F}_i$ is the linearized state-transition matrix of the error state, while $\mathbf{G}_i$ is the noise matrix of the error state, and $\boldsymbol{n}$ is the noise vector \mbox{$\boldsymbol{n}=[\boldsymbol{n}_T^T~\boldsymbol{n}_{\omega}^T~\boldsymbol{n}_{b_{\omega}}^T~\boldsymbol{n}_{a}^T~\boldsymbol{n}_{b_{a}}^T]^T$}. Using the recursive propagation method, we can get the covariance matrix $\mathbf{P}^{b_k}_{b_{k+1}}$ and Jacobian $\mathbf{J}_{k+1}$ at time $t_{k+1}$.
	
	\subsubsection{Bias Correction}
	As we propagate $\boldsymbol{\hat{\alpha}}{_{b_{k+1}}^{b_{k}}}$, $\boldsymbol{\hat{\beta}}{_{b_{k+1}}^{b_{k}}}$, $\boldsymbol{\hat{\mathcal{F}}}^{b_k}_{b_{k+1}}$ with an assumption that the bias $\boldsymbol{b}_{a_k}$, $\boldsymbol{b}_{\omega_k}$ are constant, a approximate correction will be employed when the bias changes slightly:
	\begin{equation}
	\label{eq:bias_correction}
	\begin{aligned}
	\boldsymbol{\hat{\alpha}}{_{b_{k+1}}^{b_{k}}} &\approx \boldsymbol{\hat{\alpha}}{_{b_{k+1}}^{b_{k}}} + \mathbf{J}^{{\alpha}}_{{b_{\omega}}} \delta{\boldsymbol{b}_{\omega_k}} \\
	\boldsymbol{\hat{\beta}}{_{b_{k+1}}^{b_{k}}} &\approx \boldsymbol{\hat{\beta}}{_{b_{k+1}}^{b_{k}}} + \mathbf{J}^{{\beta}}_{{b_{\omega}}} \delta{\boldsymbol{b}_{\omega_k}} \\
	\boldsymbol{\hat{\mathcal{F}}}{_{b_{k+1}}^{b_{k}}} &\approx \boldsymbol{\hat{\mathcal{F}}}{_{b_{k+1}}^{b_{k}}} + \frac{1}{\Delta t_k} \big( \mathbf{J}^{{\mathcal{F}}}_{{b_{a}}} \delta{\boldsymbol{b}_{a_k}} + \mathbf{J}^{{\mathcal{F}}}_{{b_{\omega}}} \delta{\boldsymbol{b}_{\omega_k}} \big),
	\end{aligned}
	\end{equation} 
	where $\mathbf{J}^{{\alpha}}_{{b_{\omega}}}$, $\mathbf{J}^{{\beta}}_{{b_{\omega}}}$, $\mathbf{J}^{{\mathcal{F}}}_{{b_{a}}}$, $\mathbf{J}^{{\mathcal{F}}}_{{b_{\omega}}}$ are the corresponding block in Jacobian $\mathbf{J}_{k+1}$. And $\delta{\boldsymbol{b}_{\omega_k}}$, $\delta{\boldsymbol{b}_{a_k}}$ are the bias variations between the initial estimated state and the latest estimated state in $t_k$ of gyroscope and acceleration, respectively.
	
	\subsection{Tightly Coupled VIO with External Force Estimation}
	We employ a sliding window-based tightly-coupled VIO framework as in VINS-Mono~\cite{qin2018vins} and VIMO~\cite{nisar2019vimo}.
	\subsubsection{System Formulation}
	In the sliding window, the full state of the optimization is defined as following:
	\begin{equation}
	\label{eq:fullstate}
	\begin{aligned}	
	\boldsymbol{\mathcal{X}} &=
	[
	\boldsymbol{x}_0,\boldsymbol{x}_1,\cdots,\boldsymbol{x}_n,{\lambda}_0,{\lambda}_0,\cdots,{\lambda}_m
	],  \\
	\boldsymbol{x}_k&=[\boldsymbol{p}^w_{b_k},\boldsymbol{v}^w_{b_k},\boldsymbol{q}^w_{b_k},\boldsymbol{b}_a,\boldsymbol{b}_{\omega},\boldsymbol{F}^{b_k}_{ex}],~k\in[0,n],
	\end{aligned}
	\end{equation}
where $\boldsymbol{x}_k$ is the body state at the $k$th keyframe time with the composition of position $\boldsymbol{p}^w_{b_k}$, velocity $\boldsymbol{v}^w_{b_k}$, orientation represented in Hamilton quaternion $\boldsymbol{q}^w_{b_k}$, IMU measurement bias $\boldsymbol{b}_a$, $\boldsymbol{b}_{\omega}$, and the average external force $\boldsymbol{F}^{b_k}_{ex}$ in body frame during $[t_k,t_{k+1}]$. Here, $n$ is the size of the sliding window, $m$ is the total number of features in the sliding window, and $\lambda_l$ is the inverse depth of the $l$th feature from its first observation in the sliding window as in~\cite{qin2018vins}.

In order to obtain the maximum posterior estimation of all residuals concerning $\boldsymbol{\mathcal{X}}$, we employ the nonlinear optimization problem formulation in~\cite{qin2018vins} and its extension, which considers the dynamics residual in~\cite{nisar2019vimo}. Crucially, we reformulate the dynamics residual in~\cite{nisar2019vimo} by using our proposed external force preintegration. The cost function is:
	\begin{equation}
	\begin{aligned}
	\min_{\boldsymbol{\mathcal{X}}}
	\left\{ 
	\begin{Vmatrix}
	\boldsymbol{r}{_p}
	\end{Vmatrix}^2
	+\sum_{k\in\mathcal{B}}
	\begin{Vmatrix}
	\boldsymbol{r}{_\mathcal{B}^k}
	\end{Vmatrix}^2_{\mathbf{W}{_\mathcal{B}^k}}
	+\sum_{(l,j)\in\mathcal{C}}
	\rho \big(
	\begin{Vmatrix}
	\boldsymbol{r}{_\mathcal{C}^{l,j}}
	\end{Vmatrix}^2_{\mathbf{W}{_\mathcal{C}^{l,j}}} \big) \right. \\
	\left.
	+ \underbrace{
		\sum_{k\in\mathcal{D}}
		\begin{Vmatrix}
		\boldsymbol{r}{_\mathcal{D}^k}
		\Big(\boldsymbol{x}_{k},\boldsymbol{x}_{k+1},\boldsymbol{\hat{z}}^{b_k}_{b_{k+1}}
		\Big)
		\end{Vmatrix}^2_{\mathbf{W}{_\mathcal{D}^k}}
	}_{\text{\small the external force residual}}
	\right\},
	\end{aligned}
	\end{equation}
where the first three residuals are the VIO residual terms of prior residual, the sum of the inertial residuals, and the sum of visual residuals with Huber-norm, which can be found in more details in \cite{qin2018vins}, the last one is the sum of Mahalanobis norm of our proposed external force residual, in which, $\mathcal{D}$ is the set of thrust and external force measurements in the sliding window. $\mathbf{W}{_\mathcal{D}^k}$ is the weight matrix of Mahalanobis norm for the $k$th state in the sliding window. $\boldsymbol{r}{_\mathcal{D}^k}$ is the corresponding residual at $t_k$ which considers the state $\boldsymbol{x}_{k}$, $\boldsymbol{x}_{k+1}$ and external force preintegration $\boldsymbol{\hat{z}}^{b_k}_{b_{k+1}}$.
	
	\subsubsection{External Force Residual}
	We combine $\boldsymbol{{\alpha}}{_{b_{k+1}}^{b_{k}}}$, $\boldsymbol{{\beta}}{_{b_{k+1}}^{b_{k}}}$, $\boldsymbol{{\mathcal{F}}}^{b_k}_{b_{k+1}}$, the the prediction of preintegration term $\boldsymbol{\hat{\alpha}}{_{b_{k+1}}^{b_{k}}}$, $\boldsymbol{\hat{\beta}}{_{b_{k+1}}^{b_{k}}}$, $\boldsymbol{\hat{\mathcal{F}}}^{b_k}_{b_{k+1}}$ in (\ref{eq:preintegration}), to formulate our proposed residual between frame $b_k$ and $b_{k+1}$.
	
	\begin{align}
	\boldsymbol{r}{_\mathcal{D}^k}
	\Big(\boldsymbol{x}_{k},\boldsymbol{x}_{k+1},\boldsymbol{\hat{z}}^{b_k}_{b_{k+1}}
	\Big)
	=
	\begin{bmatrix} 
	\boldsymbol{\alpha}^{b_k}_{b_{k+1}} -  \boldsymbol{\hat{\alpha}}^{b_k}_{b_{k+1}} \\
	\boldsymbol{\beta}^{b_k}_{b_{k+1}} -  \boldsymbol{\hat{\beta}}^{b_k}_{b_{k+1}} \\ 
	\boldsymbol{\mathcal{F}}^{b_k}_{b_{k+1}} -  \boldsymbol{\hat{\mathcal{F}}}^{b_k}_{b_{k+1}} \\ 
	\boldsymbol{b}_{a_{k+1}} - \boldsymbol{b}_{a_k} \\
	\end{bmatrix},\mathbf{W}{_\mathcal{D}^k}={\mathbf{P}^{b_k-1}_{b_{k+1}}},
	\end{align}
	where
	\begin{align} 
	\boldsymbol{\alpha}^{b_k}_{b_{k+1}} &= 
	\mathbf{R}{^{b_{k}}_{w}}
	\Big({{\boldsymbol{p}{_{b_{k+1}}^w}}}
	-{{\boldsymbol{p}{_{b_{k}}^w}}}
	-{{\boldsymbol{v}{_{b_{k}}^w}}}\Delta{t{_k}}
	-\frac{1}{2}\boldsymbol{\textsl{g}}^w\Delta{t{_k^2}}
	\Big) \nonumber \\
	&\quad-\frac{1}{2}\boldsymbol{F}^{b_{k+1}}_{ex}\Delta{t{_k^2}} \nonumber \\
	\boldsymbol{\beta}^{b_k}_{b_{k+1}}
	&= 
	\mathbf{R}{^{b_{k}}_{w}}
	\Big(\boldsymbol{v}{_{b_{k+1}}^w}-\boldsymbol{v}{_{b_{k}}^w}
	-\boldsymbol{\textsl{g}}^w\Delta{t{_k}}
	\Big)
	-\boldsymbol{F}^{b_{k+1}}_{ex}\Delta{t{_k}} \nonumber \\
	\boldsymbol{\mathcal{F}}^{b_k}_{b_{k+1}}
	&= 
	\boldsymbol{F}^{b_{k+1}}_{ex} .
	\end{align}

	The residual $\delta\boldsymbol{\alpha},\delta\boldsymbol{\beta},\delta\boldsymbol{\mathcal{F}},\delta\boldsymbol{b}_a$ represents the difference between the prediction model and the measurement integration of the contribution of distance by the thrust, the velocity contributed by thrust, the average force and the measurement bias deviation, respectively. $\boldsymbol{\hat{\alpha}}{_{b_{k+1}}^{b_{k}}}$, $\boldsymbol{\hat{\beta}}{_{b_{k+1}}^{b_{k}}}$, $\boldsymbol{\hat{\mathcal{F}}}^{b_k}_{b_{k+1}}$ are the corrected measurements preintegration in (\ref{eq:bias_correction}).
	Instead of a manual weight for the external force residual in VIMO,  $\mathbf{W}{_\mathcal{D}^k}$ is the weight of  Mahalanobis norm, and the covariance $\mathbf{P}^{b_k}_{b_{k+1}}$ is propagated by (\ref{eq:covarance_propagation}).
	
	As for the other part of the motion state estimation, we use the same marginalization and initialization strategy in \cite{nisar2019vimo,qin2018vins}.
	
	\subsection{Thrust Coefficient Identification}
	\label{subsec:model_identification}
Since we consider the external force acting on the center of mass and ignore the disturbance on torque, what we need to identify for the system are the mass of the drone and the thrust coefficients of the propellers.
The mass can be easily obtained, but thrust coefficient identification is often required in the force sensor, which is expensive.
Here, we employ a simple method to estimate the thrust coefficients with a given mass.
We consider the centroid and the geometry center are coincident, and each propeller from the quadrotor is equally distributed $m{\textsl{g}}/4$ gravity when the quadrotor is hovering.
Then we estimate the thrust coefficient for each propeller by the Recursive Least Squares method.
%Although the simplified model increases the system model error, it greatly benefits the usage's convenience without much damage to the estimation when the centroid and the geometry center are near.

	\section{Results}
	\label{sec:results}
	In this section, we evaluate our proposed algorithm by real-world experiments in different scenes, focusing on the accuracy and robustness of both odometry and force.
	%, where the proposed method can provide external force estimation for disturbance analysis during delivery and weighing the package as a flying scale
	Within the experiments, some comparisons among the proposed, VIMO and VINS-Mono will be presented within the same condition and the same parameters (sliding window size, noise parameters, and so on), while the weight for the external force residual in VIMO is set as the recommended one in the open-source file.
	
	\subsection{Real-world Experiments}
	
	\begin{figure}[t]
		\captionsetup{font={small}}
		\centering
		\includegraphics[width=1\columnwidth]{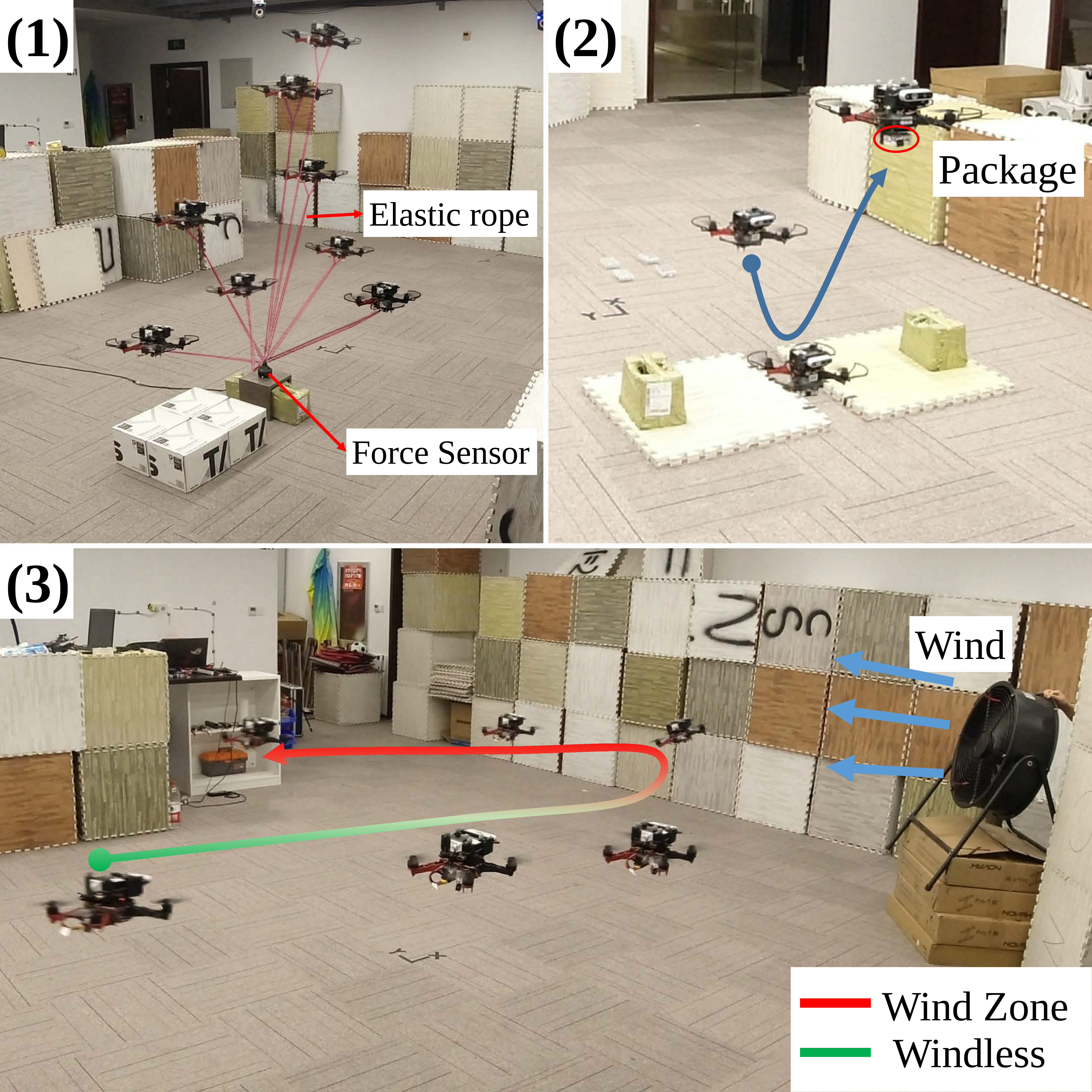}
		\captionsetup{font={small}}
		\caption{(1) shows the roped flight experiment. The drone flies with an elastic rope. (2) shows the flying scale application, the drone flies down and picks up the package, hovering for weighing. (3) shows the flight in wind, the drone is blown away in the red part.  \label{fig:3Experiments}}
		\vspace{-0.1cm}
	\end{figure}
	
	\begin{figure}[t]
		\captionsetup{font={small}}		
		\includegraphics[width=1\columnwidth]{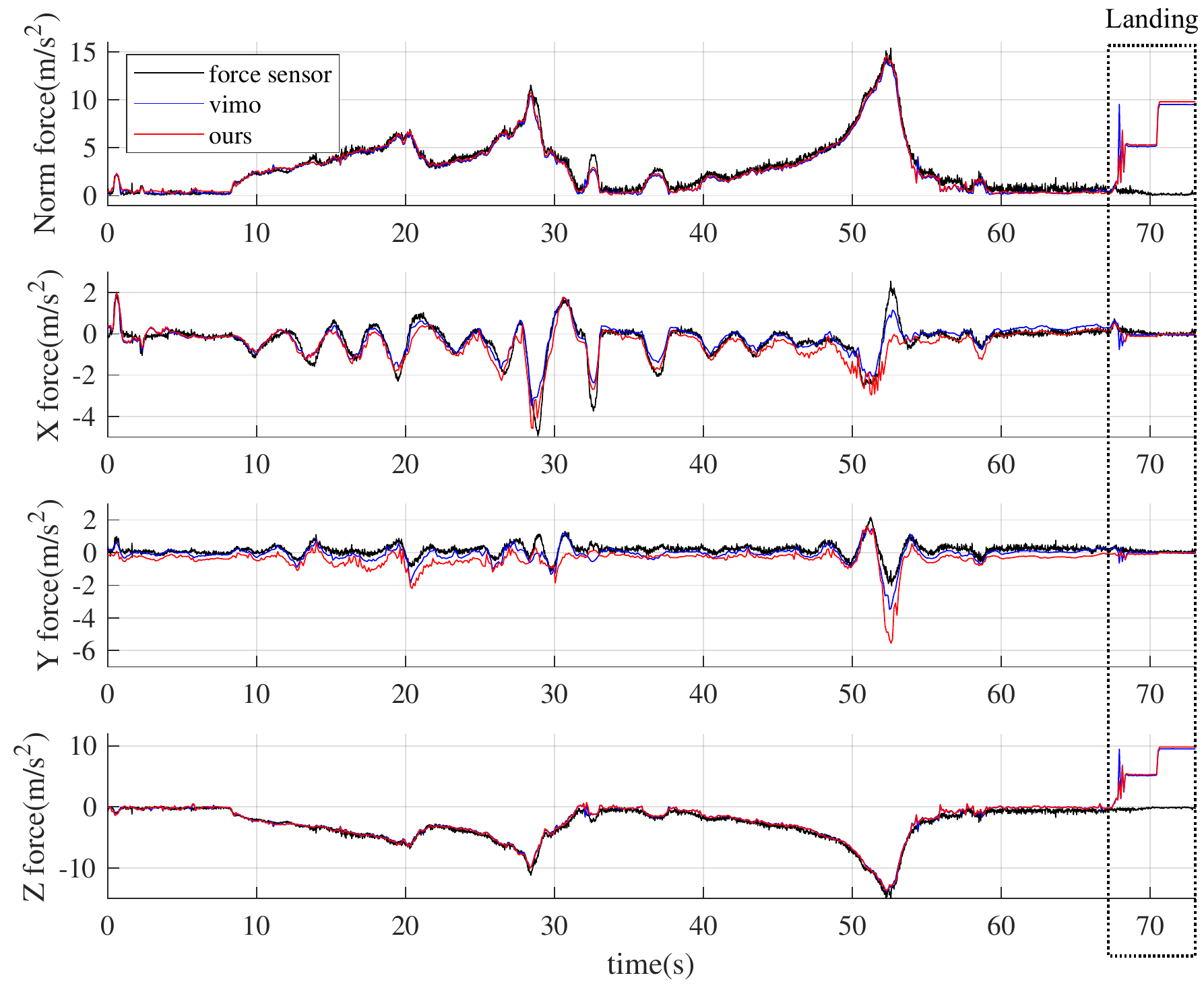}
		\caption{Force from elastic rope in norm, $x$, $y$, $z$, with RMSE $ = [0.998, 0.488, 0.751, 0.442]$. The drone lands after $66.1$s. \label{fig:force_line}}
	\vspace{-2.0cm}
	\end{figure}  

	\begin{figure}[t]
		\captionsetup{font={small}}
		\centering
		\includegraphics[width=1\columnwidth]{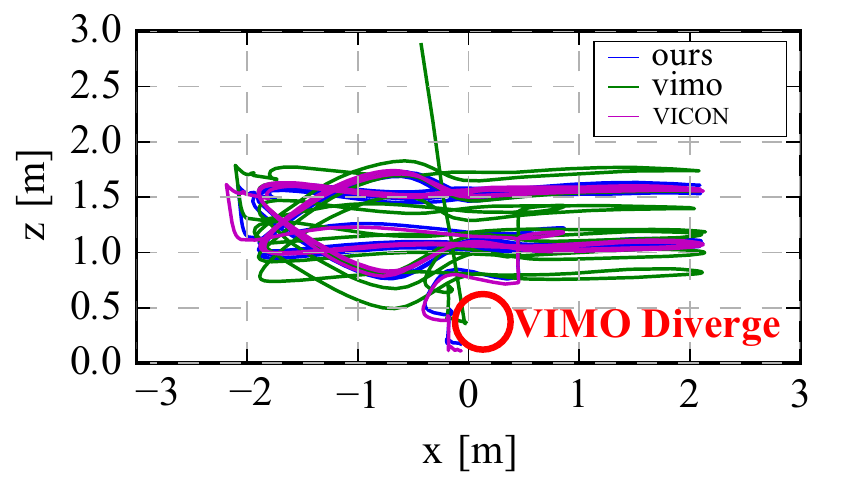}
		%\vspace{-0.5cm}
		\captionsetup{font={small}}
		\caption{Trajectories $2$ of flight with loads. The odometry estimation of VIMO is affected by the external force from loads. When the drone lands, the external force is the normal force, causing the divergence of VIMO, as illustrated above. \label{fig:VIMO_fail}}
		 \vspace{-0.1cm}
	\end{figure}
	
	\begin{figure}[t]
		\captionsetup{font={small}}
		\centering
		\includegraphics[width=1\columnwidth]{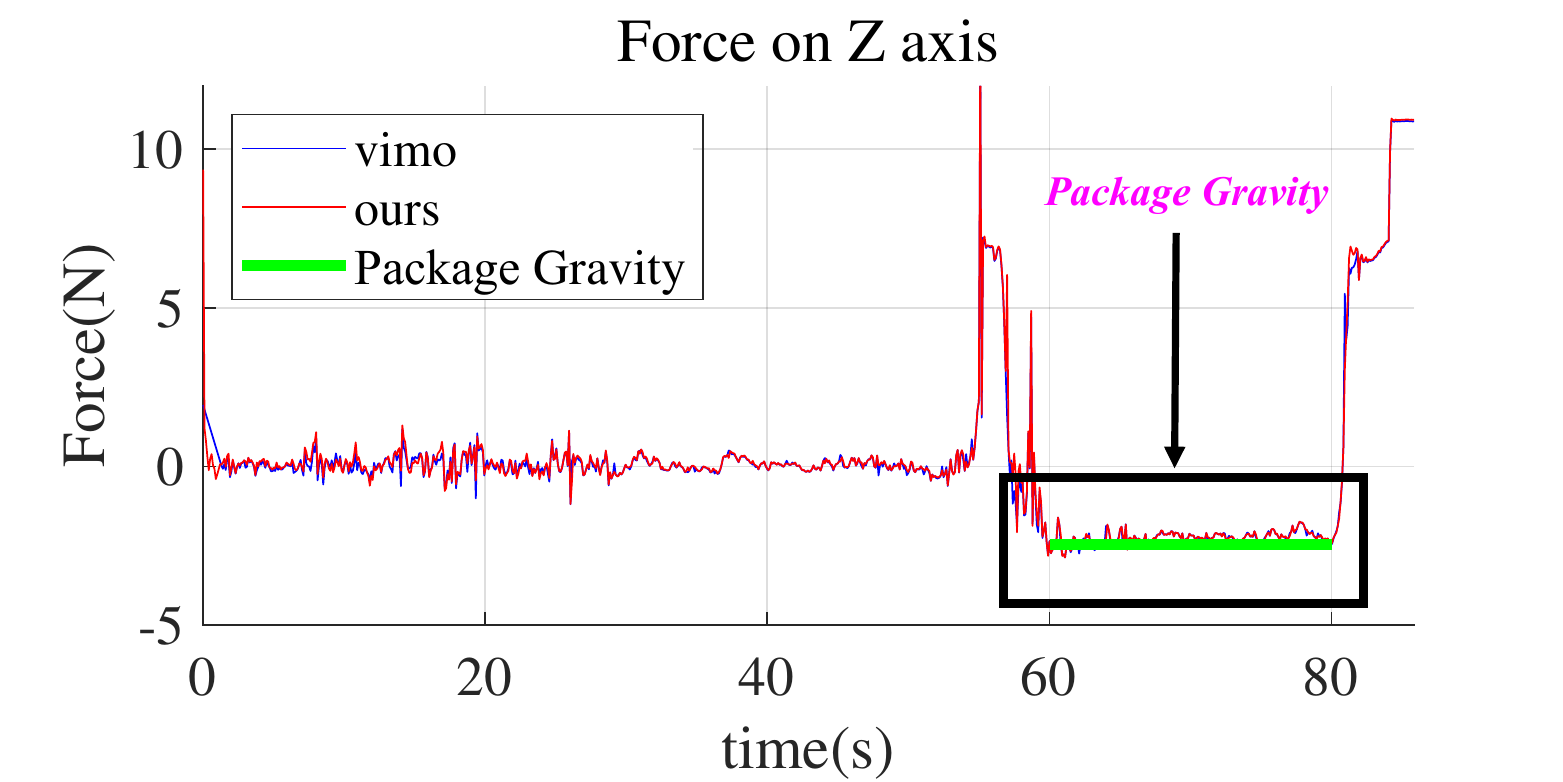}
		%\vspace{-0.5cm}
		\captionsetup{font={small}}
		\caption{Force estimation in flying scale experiment. The drone weighs the package during $60$s and $80$s. The peak before weighing is caused by the landing for the package. The RMSE between the gravity of the package and the estimation is about $0.29$N.\label{fig:flying_scale}}
		\vspace{-1.9cm}
	\end{figure}

	\subsubsection{Hardware Setup}
	As for validation in real-world experiments, we use the quadrotor platform shown in Fig.~\ref{fig:axis}, with an onboard computer (i7-8550U), a stereo camera (Intel Realsense D435), a DJI N3 controller providing IMU data.
	As for the homemade RMU, we measure the changing of rotor phases by a voltage comparator and an ARM-based micro computing unit (MCU), and then we calculate the rotor rpm by fast Fourier transform (FFT)~\cite{nussbaumer1981fast}.
	We obtain the IMU data at $400$Hz, imagery data at $30$Hz, RMU data at $100$Hz.
	We identify the thrust coefficient by the proposed method in \ref{subsec:model_identification} with a given mass of the drone.

% 	\begin{figure}[t]
% 		\captionsetup{font={small}}
% 		\centering
% 		\subfigure[Top view of flight with loads trajectory $2$]{
% 			\label{fig:subfig:tv} %% label for second subfigure
% 			\includegraphics[width=1\columnwidth]{D1027_traj8_weight_trajectory_top.pdf}}
% 		\vspace{-0.20cm}
		
% 		\centering
% 		\subfigure[Side view of flight with loads trajectory $2$]{
% 			\label{fig:subfig:sv} %% label for first subfigure
% 			\includegraphics[width=1\columnwidth]{D1027_traj8_weight_trajectory_side.pdf}}
% %		\vspace{-0.25cm}
% 		\caption{ View of flight with loads trajectory. \label{fig:flight_with_loads_trajectory}}
% 	\end{figure}

		\begin{figure}[t]
			\captionsetup{font={small}}
			\centering
			\includegraphics[width=1\columnwidth]{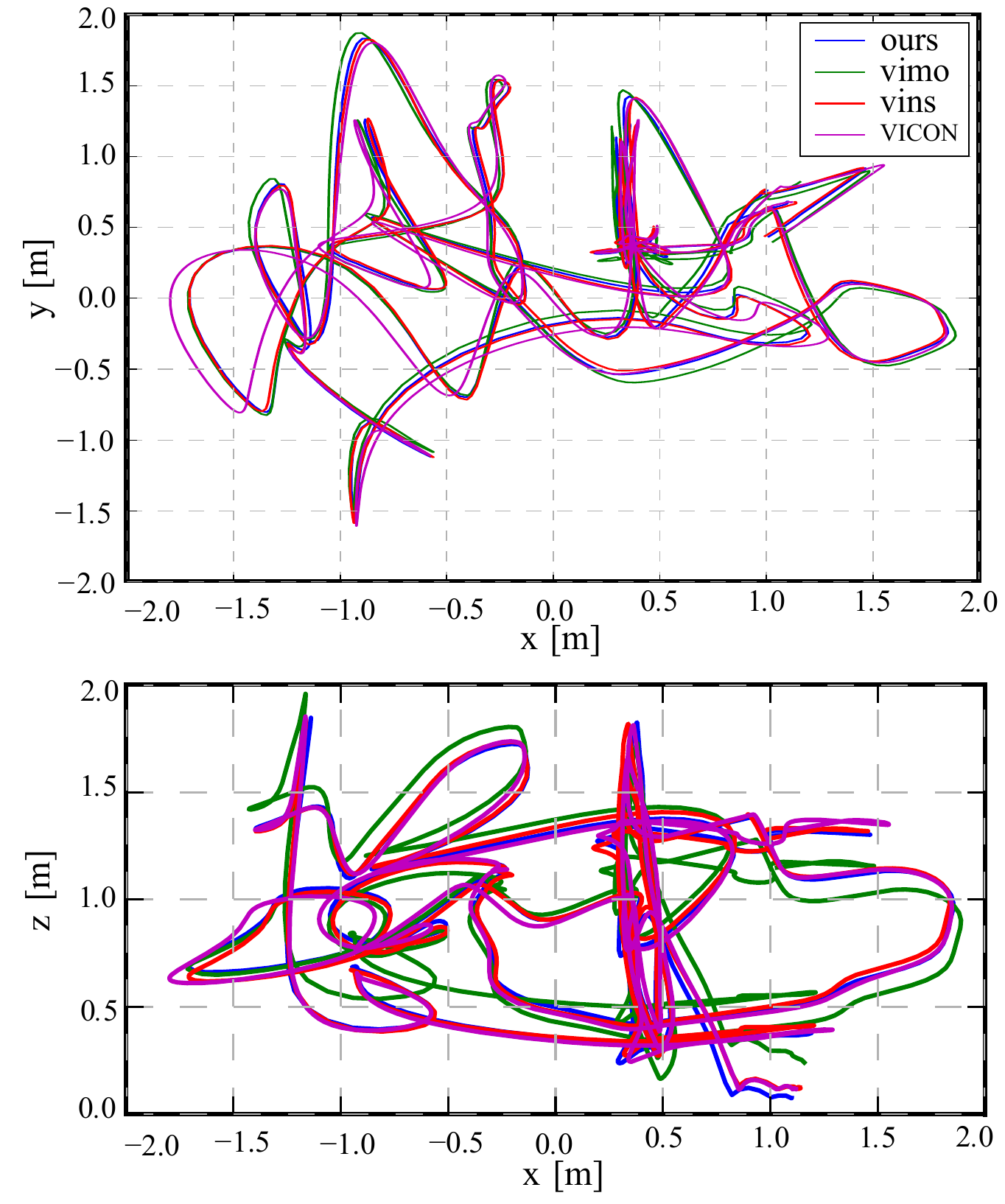}
			\vspace{-0.5cm}
			\captionsetup{font={small}}
			\caption{Trajectories $1$ of flight with loads. Ours keeps the accuracy compared with VINS-Mono, while VIMO is strongly affected by the external force, especially $z$ axis. \label{fig:flight_with_loads_trajectory}}
			\vspace{-0.2cm}
		\end{figure}
		
		\begin{figure}[t]
			\captionsetup{font={small}}
			\centering
			\includegraphics[width=1\columnwidth]{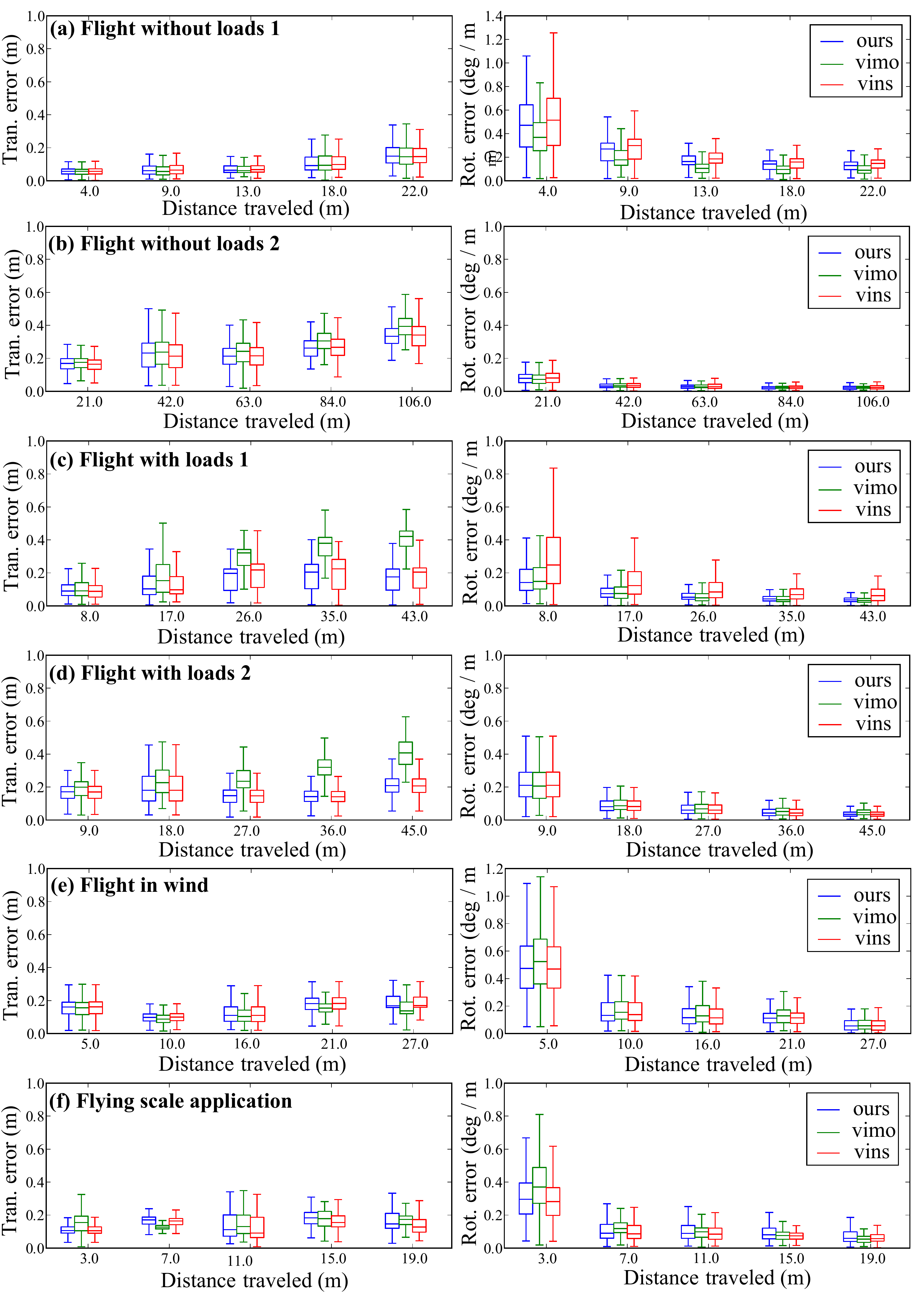}
			\vspace{-0.3cm}
			\captionsetup{font={small}}
			\caption{RMSE of translation and rotation in the experiments.   \label{fig:RMSE_traj}}
			\vspace{-0.5cm}
		\end{figure}

	\begin{table}[t]   
		\captionsetup{font={small}}
		\centering  
		%\fontsize{6.5}{8}\selectfont  
		\begin{threeparttable}  
			\caption{Real-world Experiment Evaluation}  
			\label{tab:Experiment_Evaluation}  
			\begin{tabular}{cccccccc}  
				\toprule  
				\multirow{2}{*}{Trajectory}&
				\multicolumn{3}{c}{\bf trans. RMSE(m)}&\multicolumn{3}{c}{\bf rot. RMSE(deg)} \cr  
				\cmidrule(lr){2-4} \cmidrule(lr){5-7}  
				&VINS&VIMO&Ours&VINS&VIMO&Ours&\cr  
				\midrule  
				1  w/o loads  & 0.058     & 0.058     & \bf 0.057 & 4.03     & \bf 3.36  & 3.86  \cr
				2  w/o loads  & 0.201     & 0.240     & \bf 0.198 & 4.43     & \bf 4.32  & 4.84  \cr
				1 w/ loads    & 0.132     & 0.285     & \bf 0.120 & 3.37     & 4.09      & \bf 3.22  \cr
				2 w/ loads    & \bf 0.102 & 0.239     & 0.103     & \bf 4.28 & 4.40      & \bf 4.28  \cr
				In wind       & 0.091     & \bf 0.076 & 0.090     & \bf 3.81 & 3.90      & 3.89  \cr
				Flying scale  & \bf 0.069 & 0.091     & 0.088     & 4.44     & \bf 4.28  & 4.58  \cr
				\bottomrule  
			\end{tabular}  
		\end{threeparttable} 
		\vspace{-0.4cm}
	\end{table}

\subsubsection{Trajectory Flight}
We record the data from three flight trajectories: flight without loads, flight with loads fixed on the drone, and flight in wind. 
The external force of flight without loads is the aerodynamic drag force. 
It is a combined force from loads and aerodynamic drag in flight with loads trajectories. 
The last one tests in a scene where the drone tries to fly in front of a fan, and it is finally blown away by the wind, as illustrated in Fig.~\ref{fig:3Experiments} (3), and the estimated external force will be shown in the proposed video.
As shown in Fig.~\ref{fig:VIMO_fail} and Fig.~\ref{fig:flight_with_loads_trajectory}, the proposed approach is more robust in odometry estimation when flight with loads.

To be mentioned, VIMO usually fails when the force is unneglectable large, especially in the landing period, as shown in Fig.~\ref{fig:VIMO_fail}, because the large force violates the zero-means force assumption, and the cost of the optimization cost function becomes abnormally large.
The weight of the external force term in VIMO should be adjusted as the force's magnitude.
Therefore, results shown in  Fig.~\ref{fig:flight_with_loads_trajectory} $\sim$ \ref{fig:RMSE_traj} and Tab.~\ref{tab:Experiment_Evaluation} are the segments that VIMO does not diverge, for better comparing the robustness and accuracy of the proposed. 

\subsubsection{Roped Flight}
To evaluate the estimated force's accuracy, we perform a roped flight test, connecting the drone and force sensor by an elastic rope, as shown in Fig.~\ref{fig:3Experiments} (1). The force estimation comparison is illustrated in Fig.~\ref{fig:force_line}. The result shows that our proposed method can provide accurate force estimation whenever the force is slight or large. 

\subsubsection{Application}
We apply our proposed method in a simulated scene of package delivery as an application of weighing package in the air, just like a flying scale, as shown in Fig.~\ref{fig:3Experiments} (3). The force estimation shown in Fig.~\ref{fig:flying_scale} represents the accurate force estimation performance of the proposed.

\subsection{Discussion}
As shown in Fig.~\ref{fig:RMSE_traj} and Tab.~\ref{tab:Experiment_Evaluation}, our proposed method shows similar even better performance than VINS-Mono in pose estimation, even though it performs slightly worse than VINS-Mono in some scenarios due to the noisy thrust measurements. 
In Tab.~\ref{tab:Experiment_Evaluation}, VIMO shows similar or better performance in some indexes and performs superiorly in force estimation shown in Fig.~\ref{fig:force_line} and Fig.~\ref{fig:flying_scale}, but it performs much worse in pose estimation under a large external force. 
These experiments highlight that our proposed VID-Fusion performs robustly and accurately on both odometry and external force estimation, even the external force ranges from neglectable to significant.

	\section{Conclusion}
	\label{sec:conclusion}
	This paper proposes an approach that offers robust and accurate odometry for external force estimation based on tightly-coupled optimization-based VIO framework~\cite{qin2018vins}.
	Our proposed estimator increases the robustness in odometry estimation and keeps accurate external force and pose estimation, whenever the force is large or lasts for a long time.
	
	The proposed visual-inertial-dynamics estimator framework can be easily extended into a stereo visual-inertial system such as VINS-Fusion~\cite{qin2019general}, and it can run without model measurements, going back to a VIO system, and the external force will be the combination of normal force and acceleration.
	Moreover, compared with VINS-Mono~\cite{qin2018vins}, we keep the accuracy of odometry estimation and even provide an increase in some scenarios, with an accurate external force estimation additionally.
	Since we use a simplified dynamics model in this paper, the thrust torque is not yet utilized and is left as our future work.
	Finally, as a return to the community, the source code and our customized datasets will be released\footnote{Code, and the dataset will be released after the acceptance of this paper at \url{https://github.com/ZJU-FAST-Lab/VID-Fusion}}.
	
	\bibliography{main}
\end{document}